\documentclass[sigconf]{acmart}

\AtBeginDocument{%
  }

\usepackage{multirow}
\usepackage{algorithm,algorithmic}
\usepackage{subfigure}

\usepackage{makecell}
\newcolumntype{C}[1]{>{\centering\arraybackslash}p{#1}}
\newcolumntype{P}[1]{>{\arraybackslash}p{#1}}
\usepackage{colortbl}
\usepackage{pifont}       
\usepackage{bbding}       
\usepackage{fontawesome}  
\usepackage{multirow}

\begin{document}

\title[USB-Rec: An Effective Framework for Improving Conversational Recommendation Capability of LLM]{USB-Rec: An Effective Framework for Improving Conversational Recommendation Capability of Large Language Model}

\author{Jianyu Wen}
\email{jianyuwen2000@gmail.com}
\affiliation{%
  \institution{Harbin Institute of Technology}
  \city{Shenzhen}
  \state{Guangdong}
  \country{China}
}

\author{Jingyun	Wang}
\affiliation{%
  \institution{Beihang University}
  \city{Beijing}
  \country{China}}
\email{wangjingyun0730@gmail.com}
\authornote{Equal contribution}

\author{Cilin Yan}
\affiliation{%
  \institution{Xiaohongshu Inc.}
  \city{Beijing}
  \country{China}}
\email{clyanhh@gmail.com}

\author{Jiayin Cai}
\affiliation{%
  \institution{Xiaohongshu Inc.}
  \city{Beijing}
  \country{China}}
\email{caijy18@tsinghua.org.cn}

\author{Xiaolong Jiang}
\affiliation{%
  \institution{Xiaohongshu Inc.}
  \city{Beijing}
  \country{China}}
\email{laige@xiaohongshu.com}

\author{Ying Zhang}
\affiliation{%
  \institution{Harbin Institute of Technology}
  \city{Shenzhen}
  \state{Guangdong}
  \country{China}}
\email{zhangyinghit@hit.edu.cn}
\authornote{Corresponding author}

\begin{abstract}
Recently, Large Language Models (LLMs) have been widely employed in Conversational Recommender Systems (CRSs).
Unlike traditional language model approaches that focus on training, all existing LLMs-based approaches are mainly centered around how to leverage the summarization and analysis capabilities of LLMs while ignoring the issue of training.
Therefore, in this work, we propose an integrated training-inference framework, User-Simulator-Based framework (USB-Rec), for improving the performance of LLMs in conversational recommendation at the model level.
Firstly, we design a LLM-based Preference Optimization (PO) dataset construction strategy for RL training, which helps the LLMs understand the strategies and methods in conversational recommendation.
Secondly, we propose a Self-Enhancement Strategy (SES) at the inference stage to further exploit the conversational recommendation potential obtained from RL training.
Extensive experiments on various datasets demonstrate that our method consistently outperforms previous state-of-the-art methods.
Codes are available at \url{https://github.com/John-Wendell/USB_Rec}.
\end{abstract}



\maketitle

\section{Introduction}

Conversational recommendation aims to achieve users' desired outcomes from continuous interaction between users and CRSs.
The CRS~\cite{li2018towards,sun2018conversational,zhou2020towards,jannach2021survey} continuously adjusts the recommended content according to users' feedback, thus eliciting users' preferences and making more accurate recommendations.
However, conventional CRSs~\cite{wang2023rethinking,bang-etal-2023-multitask,qin-etal-2023-chatgpt} are prone to overfitting the training set, which consequently leads to suboptimal performance on large industrial datasets.

\begin{figure}[t!]
  \centering
  \includegraphics[width=0.98\linewidth]{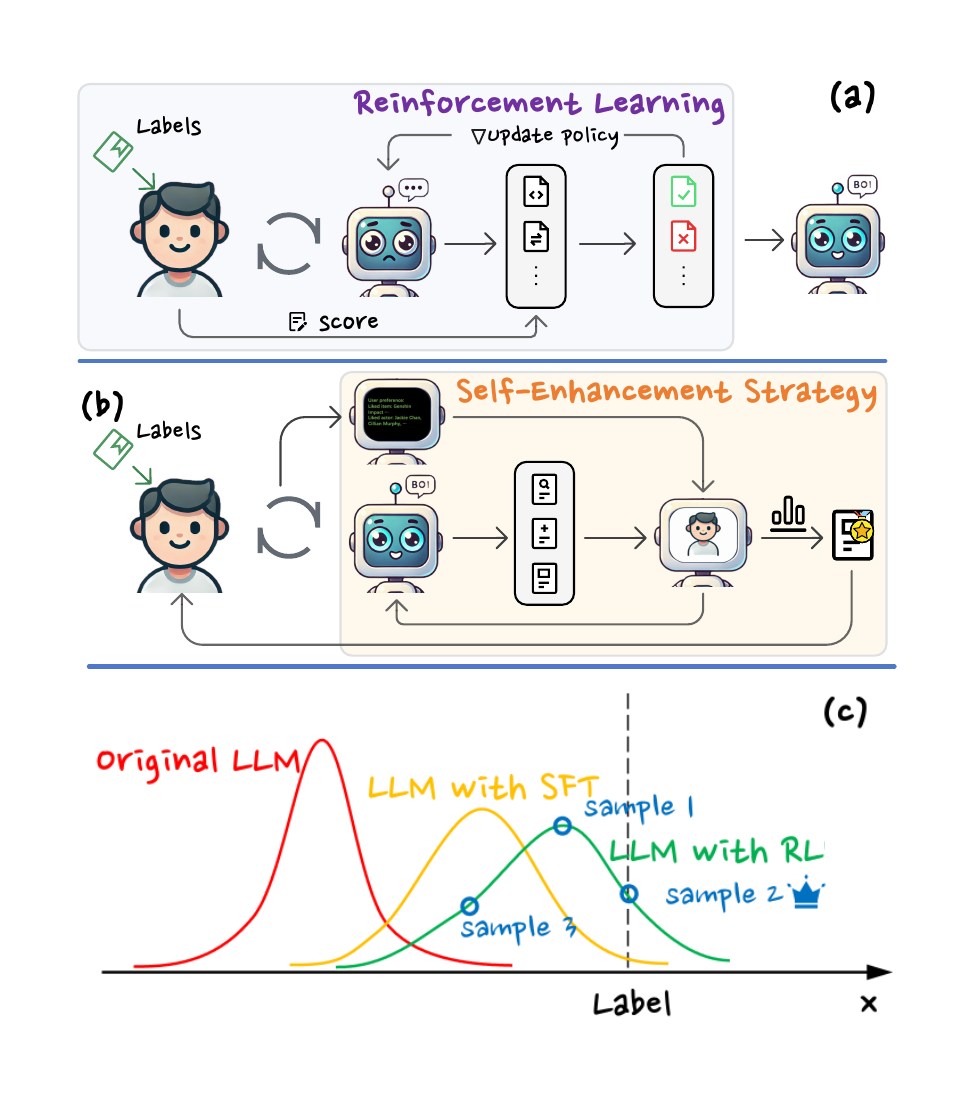}
  \caption {
  The overview of our USB-Rec framework. 
  (a) shows the reinforcement learning scheme we designed, 
  (b) is the inference search strategy we proposed,
  and (c) illustrates why the above two components are effective when combined.
  }
  \label{fig:teaser}
\end{figure}

Over the past few years, Large Language Models (LLMs) have demonstrated general applicability across various domains~\cite{abbasiantaeb2024let,wu2024autogen,dubey2024llama,glm2024chatglm}.
Recent works~\cite{zhao2024recommender,chen2024large,wu2024survey} exploit LLMs as the core component of CRSs.
These methods predominantly focus on developing intricate pipelines or sophisticated prompts to optimize response quality of LLMs.
And, current research efforts have neglected training-centric approaches for enhancing LLMs intrinsic capabilities in conversational recommendation at the model level.
Endowing LLMs with intrinsic conversational recommendation capability at the model level could eliminate lots of complex pipelines and make the CRSs more streamlined and efficient.
However, fine-tuning LLMs faces substantial technical hurdles.
Supervised Fine-Tuning (SFT) methods heavily rely on the training datasets, which is prone to induce overfitting for LLMs.
In addition, conversational inconsistencies among different recommenders during data collection introduce significant noise, which adversely affects the model's learning efficacy.
While Reinforcement Learning (RL) strategies enhance LLMs generalization and ensure stylistic consistency in responses, they relies on a costly and time-consuming manual scoring process.

In this work, we propose a User-Simulator-Based framework (USB-Rec) to exploit LLMs' conversational recommendation ability through synergistic integration of model-intrinsic optimization during both training and inference stages.
Firstly, we propose an LLM-based Preference Optimization dataset construction method for RL training.
We adopt a simulated user to evaluate the responses generated by LLM and subsequently replace the original labels with those of higher scores, thus constructing a preference pair dataset for further RL training.
After training with RL based on the preference pair dataset, the output distribution of LLM tends to converge closer to the target than that achieved with SFT.
Consequently, we propose SES at the inference stage to find the most appropriate response.
Figure~\ref{fig:teaser}(a) illustrates the PODCS process in Section 3.1. A user simulator with label engages in a simulated conversation with the base recommender LLM, featuring an unhappy-faced robot icon, and scores its multi-sampled responses. These scores are regarded as reinforcement learning reward signals for RL training the recommender. The fine-tuned recommender agent is represented by the smiling-faced robot icon on the right.
Figure~\ref{fig:teaser}(b) shows the pipeline of SES framework in Section 3.2. A user preference summarizer at the top is used to construct an internal user simulator by summarizing preceding dialogues. This internal user, the icon with the user's image on the robot's screen, then is deployed to score the multi-sampled responses of the recommender LLM. The optimal response is ultimately returned to the user on the left side.
Figure~\ref{fig:teaser}(c) shows that with RL-based training on the preference pair dataset, the distribution of LLM's responses tends to converge closer to the target, which shows in Section 4.
We sample multiple responses of the CRSs and select the final response by our proposed internal interaction mechanism.


Our main contributions are summarized as follows:

(1) We design an automatic and LLM-based preference data construction strategy for RL training, which reduces the noise by using a user simulator to score the generated conversation.

(2) We propose a Self-Enhancement Strategy (SES) at the inference stage to further exploit the conversational recommendation potential obtained from RL stage.

(3) Extensive experiments on various datasets demonstrate that our method consistently outperforms previous state-of-the-art methods.

\section{Related Work}

\subsection{Traditional CRSs}
Since early natural language models lacked a large knowledge basis and enough pre-training as well as sufficient amount of parameters, they were not able to perform a series of multiple tasks such as recommending and generating replies at the same time while understanding the user's preference.
Traditional CRSs tend to separate the work of context understanding, recommending items and response generation into mutually independent parts~\cite{li2018towards,moon2019opendialkg,chen2019towards,kang2019recommendation,zhou2020improving,wang2022towards,li2022user,deng2023unified}, even though some of them utilized some early pre-trained models, like GPT2 or DialoGPT~\cite{wang2022towards,deng2023unified}.
Therefore, these methods often need to be accomplished with the support of external tools such as knowledge graphs~\cite{moon2019opendialkg,chen2019towards,zhou2020improving,wang2022towards,zhang2023variational}.

\subsection{LLM-based CRSs}
In recent years, LLMs have demonstrated their powerful capabilities for analysis, summarization, logical reasoning, and conversational communication~\cite{huang2023towards,chu-etal-2024-navigate,huang-etal-2024-queryagent,chen-etal-2024-seer,ramprasad-etal-2024-analyzing,dong-etal-2024-attacks}.
Consequently, more and more LLMs are being used in recommender systems (RSs)~\cite{liu2023llmrec,bao2023tallrec,yang2023palr,lin2024data,ji2024genrec,liao2024llara} or CRSs~\cite{friedman2023leveraging,he2023large,gao2023chat,xi2024memocrs,liang-etal-2024-llm}.
However, most of the works on LLM-based CRS tend to focus on how to build an effective prompt for the LLM to generate appropriate responses, \emph{e.g.}, Chat-Rec~\cite{gao2023chat} proposes to first retrieve recommended items with a RS and merge them into the LLM's prompt to get the response, and MemoCRS~\cite{xi2024memocrs} extracts and analyzes the user's history dialogues with a LLM and pass it to the prompt of the LLM to be used for generating responses.
These works ignore how to radically improve the recommendation ability of large models through training.
Wang et al.~\cite{wang-etal-2023-improving-conversational} propose a LLM-based data augmentation scheme for training CRSs, but they don't use it to fine-tune LLMs and it lacks a strategy for constructing the dataset for RL.
While ReFICR~\cite{yang2024unleashing} utilizes GRITLM~\cite{muennighoff2024generative} to retrieval related conversations and items to help the process of SFT, it remains constrained by intricate retrieval pipelines and prompt engineering requirements.
Friedman et al.~\cite{friedman2023leveraging} propose a RL training method for reinforcement learning of CRSs using human feedback~\cite{ouyang2022training}, this method is labor intensive and not easily scalable.

In addition, for CRSs with LLMs, traditional metrics are no longer applicable for evaluate the response of LLMs, so more evaluation systems based on LLMs have been proposed~\cite{wang2023rethinking,zhu2024llm,yang2024behavior,zhu2024reliable}.

Therefore, following recent LLM-based evaluation methods, we design and deploy a user simulator for building a preference dataset to fine-tune LLMs, and we further propose a Self-Enhancement mechanism to deeply exploit the conversation recommendation ability of LLMs.

\section{Methodology}

In this section, we propose a User-Simulator-Based framework (USB-Rec) for LLM-based conversational recommendation.
Firstly, we design a Preference Optimization dataset collection strategy (Sec.~\ref{sec:rl}), which leverages LLM-based simulated user agents to score generated responses to realize human-free and efficient RL training.
Secondly, we propose a Self-Enhancement Strategy (SES) at the inference stage (Sec.~\ref{sec:ses}) to further exploit the conversational recommendation potential obtained from the RL stage.

\begin{figure*}[htbp]
\centering
  \includegraphics[width=0.999\linewidth]{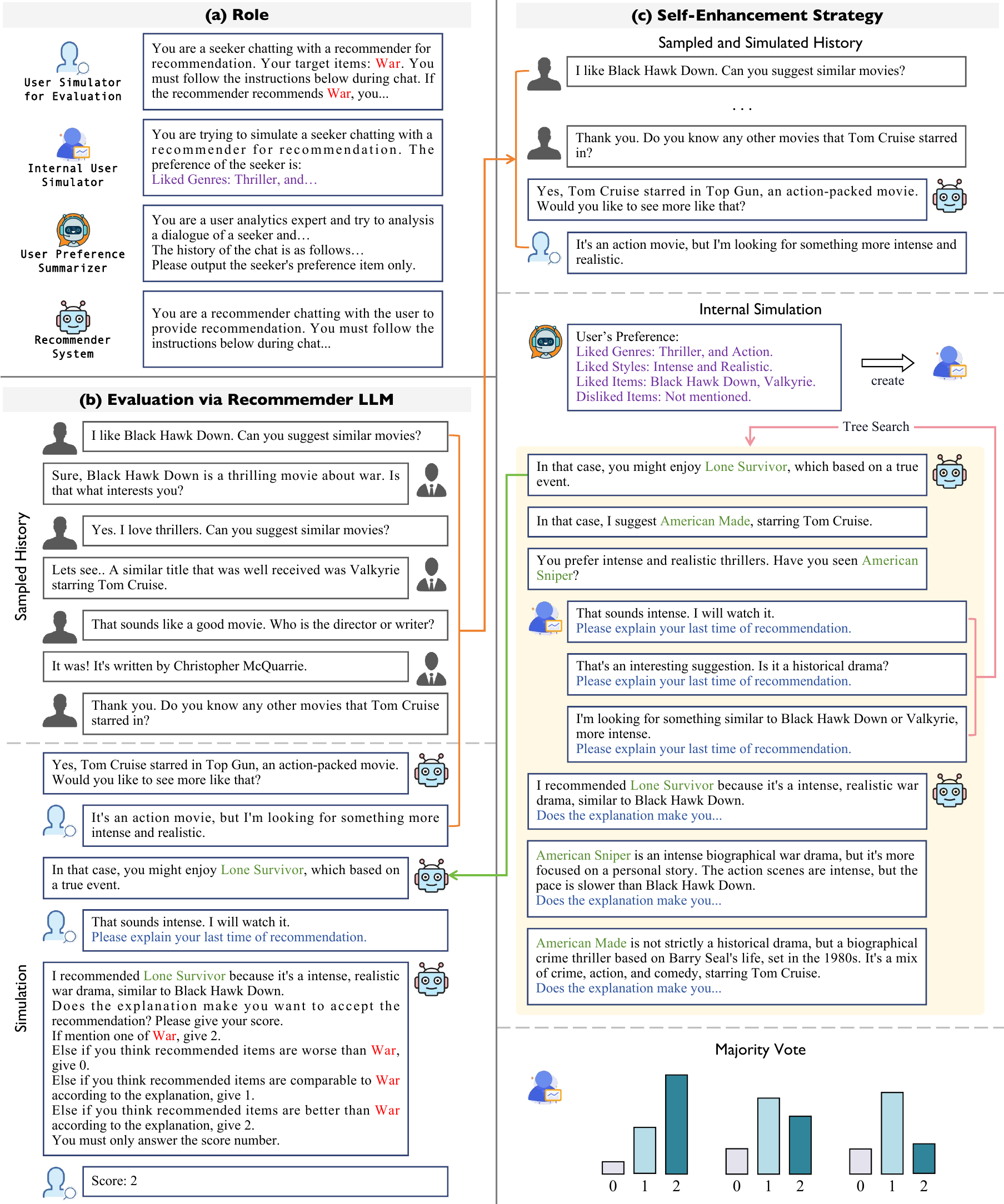}
  \caption{Overview of SES with an example. The red words are the label of the sampled turn, the green words are the recommended items given by Recommender System, the purple words are the user's profile summarized by User and the blue words are the fixed prompts added manually for evaluation. (a) An example of the pipeline how the Recommender LLM is evaluated. (b) The workflow of our SES for RecLLM system.}
  \label{fig:example}
\end{figure*}

\subsection{PO Dataset Construction Strategy (PODCS)}\label{sec:rl}
We propose a method for scoring the responses of a recommender system by a user model to construct a high-quality preference dataset, to be used for RL optimizing LLMs.

\noindent \textbf{LLM-based User Simulation}
Following iEvaLM~\cite{wang2023rethinking}, we employ an LLM as a user simulator, which assigns scores to the generated recommendations.
Provided with labels, the user simulator engages in simulated conversations with the recommender system, utilizing sampled conversations from the dataset as seeds.
After a predetermined number of conversational rounds, the user simulator assigns a score to the recommender system as follows:
\begin{equation}\label{eq:score}	
s = 
\left\{\begin{aligned}	
		&2, \,\text{prediction} \geq \text{label}\\
		&1, \,\text{prediction} \approx \text{label}\\
            &0, \,\text{prediction} < \text{label}\\
\end{aligned}\right.\\
\end{equation}
When the recommendation of of CRS aligns with the label, or when the user simulator considers the recommended items as superior to the label, we assign a score of 2.
When the user simulator perceives the recommended items to be comparable to the label, we assign a score of 1.
However, we assign a score of 0 if the user simulator assesses the recommendation results as inferior to the label.

We utilize the user simulator for both preference dataset construction at the training stage and the evaluation at the inference stage.

\noindent
\textbf{Preference Dataset Construction}
SFT training heavily relies on the quality of training datasets.
However, existing datasets are extremely noisy, which substantially limits the performance.
Therefore, we design a new preference dataset construction scheme for conversation recommendation for RL training.

Specifically, we employ the optimal model from the SFT stage as the recommender, while the original model acts as the user in the simulated conversation. 
An LLM-based user simulator is utilized to provide feedback.
Then, as shown in Algorithm~\ref{alg:dpo_data}, we collect the preference pairs for preference dataset construction. 
For each sample from the training dataset, we execute the conversation simulator $\mathcal{S}(\theta, \phi; h_i)$ for $k$ times, with high-temperature settings. 
The conversation simulator will generate a sequence of 2n-turns, comprising the user simulator's utterances $\mathcal{U} = \{u_0, u_1,\cdot\cdot\cdot, u_n\}$, recommender's responses $\mathcal{R} = \{r_0, r_1,\cdot\cdot\cdot, r_n\}$, and the final score $s$.
We select responses with high scores as acceptable completions and those with low scores as rejected completions.
For straightforward examples, it is common to reach a situation with all scores as 2.
We randomly select one as preferred data, while treating the original labels in the dataset as dispreferred data.
However, numerous challenging samples remain suboptimal or receive low scores even after undergoing SFT, which may adversely affect subsequent RL training.
Therefore, we set the preferred completion as ground truth, while the dispreferred completion as the low-scored response.
To make the preference pair reliable, we adopt a majority voting strategy.
Finally, $r_w$ of the preference pair dataset could be expressed as
\begin{equation}\label{eq:rw}	
r_w = 
\left\{\begin{aligned}	
		&r \hspace{1mm} \text{with }\hspace{1mm}s=2, \exists s=2 \hspace{1mm} \\
		& \text{label}, \forall s<2
\end{aligned}\right.\\
\end{equation}
and the $r_l$ could be written as
\begin{equation}\label{eq:rl}	
r_l = 
\left\{\begin{aligned}	
		&r \hspace{1mm}  \text{with} \hspace{1mm}s<2, \forall s<2 \\
		& \text{label}, \forall s=2
\end{aligned}\right.\\
\end{equation}

\begin{algorithm}[!ht]
    \renewcommand{\algorithmicrequire}{\textbf{Input:}}
	\renewcommand{\algorithmicensure}{\textbf{Output:}}
	\caption{Preference Dataset Construction Pipeline:}
    \label{alg:dpo_data}
    \begin{algorithmic}[1] 
        \REQUIRE  Parameters of user simulator model $\theta$, parameters of recommender system model $\phi$, the dialogue history $h$ of training dataset $\mathcal{D}$, conversation simulator $\mathcal{S}(\theta, \phi; \mathcal{D}_i)$, and Majority Voting process $\mathcal{M}$ for score $s$; 
        
        \STATE $\mathcal{D}_{pref} \gets []$
        \FORALL {$i = 0, 1, \cdot\cdot\cdot,$  length($\mathcal{D}$)}
            \STATE $h_i \gets$ the dialogue history of $\mathcal{D}_i$
            \STATE $r_{w}, r_{l} = r_{gt}, r_{gt}$;
            \FORALL {$j = 0, 1, \cdot\cdot\cdot, k$}
                \STATE $s_j, \mathcal{U}_j, \mathcal{R}_j \gets \mathcal{M}(\mathcal{S}(\theta, \phi; h_i))$;

                \IF {$ s_j == 2$}
                    \STATE $r_{w} = \mathcal{R}_j^0$;
                \ELSIF {$ s_j < 2$}
                    \STATE $r_{l} = \mathcal{R}_j^0$;
                \ENDIF
            \ENDFOR

            \STATE $\mathcal{D}_{pref} \cup \{r_{w},  r_{l}\}$;
        \ENDFOR
        
	\ENSURE $\mathcal{D}_{pref}$; 
     
    \end{algorithmic}
\end{algorithm}



\subsection{Self-Enhancement Strategy (SES)}~\label{sec:ses}
After the training stage, the LLM is equipped with the potential for multi-round conversation recommendation, but such potential cannot be fully utilized due to the noise in the training dataset and the inherent noise in RL training.
Although the distribution of the output is close to the target, it remains dispersed.
Therefore, we proposed SES to further exploit the potential of the fine-tuned LLM for multi-round conversational recommendation.
SES induces high-quality responses from an LLM by multiple times sampling with a high-temperature setting.
With an internal scoring mechanism, SES facilitates to capture the most appropriate recommendation.

Figure~\ref{fig:example} primarily describes the workflow of SES with a sample. 
Figure~\ref{fig:example}(a) illustrates the functions of the modules used in SES, as well as their corresponding role-specific prompts.
When a certain round of conversation in Figure~\ref{fig:example}(b) is in SES, the user preference summarizer in Figure~\ref{fig:example}(c) will first summarize previous dialogues into a user preference profile. 
This profile is then used to construct an internal user simulator. 
The yellow-highlighted section in Figure~\ref{fig:example}(c) represents internal simulated conversation: The recommender LLM samples multiple responses to the user's current utterance in Figure~\ref{fig:example}(b), each of which subsequently engages in extended simulated conversation with the internal user simulator until termination, with their final scores obtained through the majority voting. 
The tree search strategy further extends this process by having the recommender LLM iteratively sample multiple responses at each turn of the internal simulated conversation, each branching into new simulated dialogues with the internal user simulator. 
Ultimately, the CRS returns the response from the initial sampling batch whose sub-nodes achieve the highest aggregated scores to the user simulator for evaluation in Figure~\ref{fig:example}(b).

\noindent
\textbf{User Preference Summarizer}
Aiming to assign a score to the responses sampled from LLM, we design an internal user model to filter the response preferences.
Though without access to labels, the internal user simulator model is expected to act as closely as possible with the external user simulator model, which has label access, thus filtering consistent response preferences with the external user simulator model.
Therefore, it is essential to capture an accurate and functional user profile.
Instead of simply passing the user's historical conversation information to the internal user simulator, we leverage an LLM, which summarizes the user preferences from historical conversations and generates a fixed-format user profile.
The user profile is then fed into the internal user model along with the historical conversations as a prompt.
The expression of the user preference summarizer is
\begin{equation}
  \label{eq:summarizer}
  p = \mathcal{F}^{ups}(h^e),
\end{equation}
where the $h^e$ means the conversation history in the external evaluation process, $\mathcal{F}^{ups}$ is the LLM function with a fixed input prompt template, and $p$ is the user profile.

\noindent
\textbf{Internal User Simulator}
With the user's preference profile and historical conversation information, the internal user simulator assigns scores to the recommendation and explanation of the recommender system after a specific number of rounds.
The internal user's response process can be expressed as
\begin{equation}
  \label{eq:internal_user}
  r_i = \mathcal{F}^{ius}(h^s,p),
\end{equation}
where
\begin{equation}
  \label{eq:history}
  h^s = h^e \parallel h^i.
\end{equation}
where $h^i$ is the dialogue history between the recommender system and the internal user simulator, $\parallel$ represents the concatenation operation, $r_i$ means the response of the internal user simulator in the i-th round in within the simulated conversation, and $\mathcal{F}^{ius}$ is function of internal user simulator LLM with a fixed prompt template.

\noindent
\textbf{Search Strategy}
To explore the potential of fine-tuned LLMs' multi-round conversational recommendation, we design a search strategy to identify the best possible responses from diverse outcomes. 
First, we increase the temperature of the LLM at each round of interaction with the external user simulator to encourage the generation of more diverse responses. 
Then, each sample is scored using a tree search strategy combined with majority voting, based on the independent interactions between the internal user and the recommender system.


Tree search strategy is designed to to mitigate the bias in scoring due to the internal simulator's lack of awareness of true labels. 
Specifically, with the tree search strategy, SES iteratively samples multiple responses at each turn.
Each sampled response further branches into a new simulated dialogue with the internal user simulator.
Ultimately, the CRS returns the initial response whose sub-nodes achieve the highest aggregated scores.
As a result, the tree search strategy mitigates the potential bias in scoring.
Thanks to this search strategy and majority voting, the quality of the scoring is ensured.

\section{Experiments}
In this section, we first present the details of the experimental setup, then provide a comparison with other methods, and finally analyze the effect of the each components on the performance.





\subsection{Experimental Settings}

\begin{table}[t]
\caption{The comparison of our and previous state-of-the-art methods on ReDial and OpenDialkg datasets.}
\begin{tabular}{ccccc}
\toprule
\multirow{2}{*}{Methods}        & \multicolumn{2}{c}{iEval}  & \multicolumn{2}{c}{Recall@1}\\ \cmidrule(lr){2-3}  \cmidrule(lr){4-5} 
                                & ReDial & OpenDialkg & ReDial & OpenDialkg  \\
\midrule
KBRD~\cite{chen2019towards} & 0.79   & 0.91 & 0.028   & 0.231\\
BARCOR~\cite{wang2022barcor} & 0.82   & 1.22  & 0.031   & \textbf{0.312}\\
UniCRS~\cite{wang2022towards} & 1.08   & 1.30   & \underline{0.050}   & \underline{0.308}    \\
\midrule \noalign{\vskip 0.5ex}
GPT-3.5-turbo &  1.15  & 1.29   & 0.037   & 0.156    \\
GPT-4~\cite{achiam2023gpt} & 1.20   & \underline{1.33}   & 0.043   &0.277    \\
ReFICR~\cite{yang2024unleashing} & \underline{1.25} & -  & \textbf{0.056}& - \\
\midrule \noalign{\vskip 0.5ex}
 \rowcolor{gray!20} USB-Rec & \textbf{1.29}   & \textbf{1.40}     & \underline{0.050}   & 0.300   \\  
\bottomrule
\end{tabular}

\label{tab:compare}
\end{table}

\begin{table*}
\caption{\label{tab:two_dataset_comp_3llm}
The performance of LLMs with SFT, RL (SimPO) and SES, on two popular conversation recommendation dataset. The simulated number of rounds is set to 3 and SES is used only in the last round.
}
\centering
\begin{tabular}{cccccccc}
\toprule
\multirow{2}{*}{Models}  & \multirow{2}{*}{Datasets} & \multicolumn{6}{c}{Methods}                      \\ \cmidrule(lr){3-8} 
                         &                           & B/L  & SFT  & RL & SES& SFT+SES&  \cellcolor{gray!20}RL+SES\\ \midrule
\multirow{3}{*}{Llama3.1-8B}& ReDial                    & 1.18 (-)& 1.22 (+0.04)& 1.23 (+0.05)& 1.25 (+0.07)& \underline{1.26 (+0.08)}&  \cellcolor{gray!20}\textbf{1.29 (+0.11)}\\  
                         & OpenDialkg                & 1.28 (-) & 1.29 (+0.01)& 1.30 (+0.02)& 1.38 (+0.10)& \underline{1.39 (+0.11)}&  \cellcolor{gray!20}\textbf{1.40 (+0.12)}\\ 
                         & Average & 1.23 (-) & 1.26 (+0.03) & 1.27 (+0.04) & 1.32 (+0.09)  & \underline{1.33 (+0.10)}  &  \cellcolor{gray!20}\textbf{1.35 (+0.12)} \\ \midrule
\multirow{3}{*}{ChatGLM3-6B}& ReDial                    & 1.03 (-)& 1.05 (+0.02)& 1.06 (+0.03)& 1.08 (+0.05)& \underline{1.12 (+0.09)}&  \cellcolor{gray!20}\textbf{1.13 (+0.10)}\\  
                         & OpenDialkg                & 1.09 (-)& 1.11 (+0.02)& 1.12 (+0.03)& 1.14 (+0.05)& \underline{1.19 (+0.10)}&  \cellcolor{gray!20}\textbf{1.20 (+0.11)}\\   
                         & Average & 1.06 (-)& 1.08 (+0.02)& 1.09 (+0.03)& 1.11 (+0.05)& \underline{1.16 (+0.10)}&  \cellcolor{gray!20}\textbf{1.17 (+0.11)}\\  \midrule
\multirow{3}{*}{Qwen2.5-7B}& ReDial                    &  0.97 (-)& 1.00 (+0.03)& 1.02 (+0.05)& 1.01 (+0.04)& \underline{1.05 (+0.08)}&  \cellcolor{gray!20}\textbf{1.09 (+0.12)}\\  
                         & OpenDialkg                & 1.17 (-)& 1.19 (+0.02)& 1.20 (+0.03)& 1.19 (+0.02)& \underline{1.27 (+0.10)}&  \cellcolor{gray!20}\textbf{1.29 (+0.12)}\\   
                         & Average & 1.07 (-)& 1.10 (+0.03)& 1.11 (+0.04)& 1.10 (+0.03)& \underline{1.16 (+0.09)}&  \cellcolor{gray!20}\textbf{1.19 (+0.12)}\\  
\bottomrule
\end{tabular}

\end{table*}

To demonstrate the rationality, effectiveness, and generalizability of our approach, we test several open-source LLMs on two public CRS datasets, and study each parameter scrutinarily.

\noindent
\textbf{Datasets} 
We experiment on two popular real-world conversational recommendation datasets: 
1) ReDial~\cite{li2018towards} is a conversational movie recommendation dataset created by 504 crowd-sourcing workers on Amazon Mechanical Turk (AMT) with 10,006 dialogues and 182,150 utterances, including 51,699 related movies. 
2) OpenDialkg~\cite{moon2019opendialkg} is a cross-domain conversational recommendation dataset with 6,429 dialogues about movies and 5,891 dialogues about books, including 37,838 and 34,035 utterances respectively. We follow iEvaLM~\cite{wang2023rethinking} to split the training, validation, and testing sub-datasets for both datasets.

\begin{figure*}[tb]
  \centering
  \includegraphics[width=\linewidth]{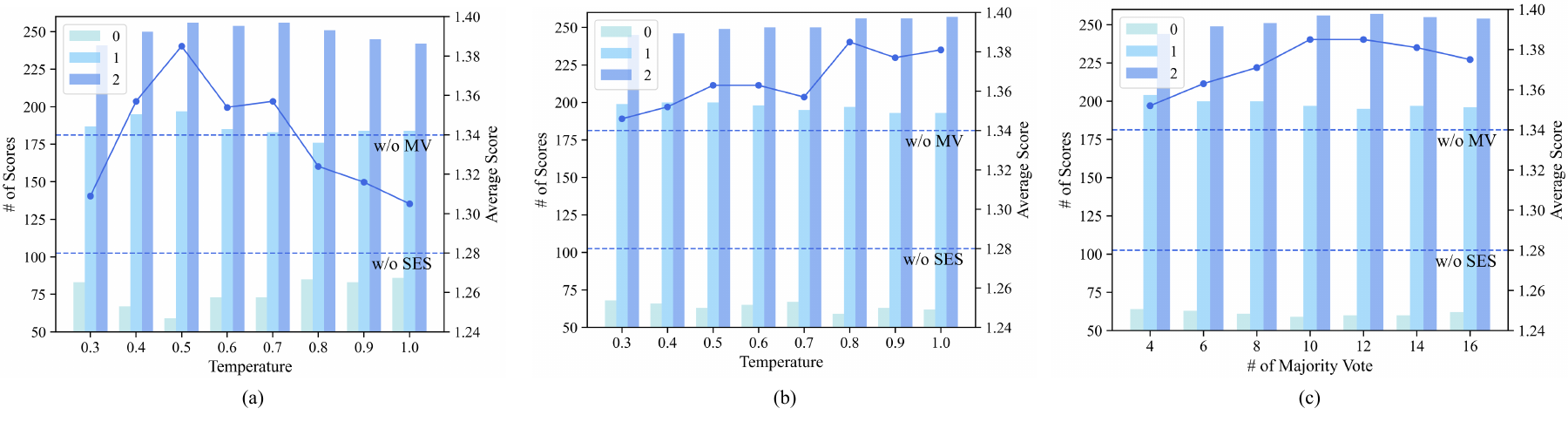}
  \caption {The ablation study of SES on OpenDialkg dataset with Llama3.1-8B. (a) shows the effect of the recommender system's temperature during the first round of sampling in SES on the final scoring. (b) and (c) show the effect of the temperature and number of majority vote on the final scores, respectively.}
  \label{fig:ablation}
\end{figure*}

\noindent
\textbf{Implementation Details} 
For both SFT and PO stages, we deploy LoRA to finetune the LLMs based on the LLaMa-Factory~\cite{zheng2024llamafactory} platform. 
We employ LoRA fine-tuning to reduce the trainable parameters. 
We set the rank of LoRA to 8, batch size to 128, and search learning rate from 1e-5 to 1e-8.
On the SFT stage, we set loss to cross-entropy and learning rate to 2e-7 and 5e-7 for ReDial and OpenDialkg dataset, respectively. 
In the RL stage, we adopt SimPO~\cite{meng2024simpo} and set $\beta$ to 0.5. 
When collecting preference data following Algorithm~\ref{alg:dpo_data}, we set the temperature of recommender LLM in the first round of $\mathcal{S}(\theta, \phi; h_i)$ to 0.5. 
What's more, since the final scoring is discrete, even if the temperature is tuned up, it is infeasible to avoid a large number of score repetitions for the same sample, especially when confronting extreme examples such as the extremely difficult or simple ones, while the time and computational consumption of sampling is high, so we chose $k = 2$.

\noindent
\textbf{Baselines} 
In selecting the comparative baselines, we considered approaches encompassing both traditional NLP-model-based CRSs and LLM-based CRSs.
Tradition NLP-model-based CRSs include KBRD~\cite{chen2019towards}, BARCOR~\cite{wang2022barcor} and UniCRS~\cite{wang2022towards}.
For LLM-based CRSs, we choose GPT-3.5-turbo and GPT-4~\cite{achiam2023gpt}, two prominent conversational LLMs, with the same prompt of our method.
Besides, we also select the ReFICR~\cite{yang2024unleashing} method, an GRITLM-based CRS which fine-tuned on conversational recommendation task, to the comparison of LLM-based CRSs.

\noindent
\textbf{Evaluation Metrics} 
Traditional recommendation evaluation metrics, like Recall@1, is proved that they tend to lead to unfair evaluation for CRS with LLM to a strong preference for the traditional recommender systems that overfit the items in the datasets~\cite{wang2023rethinking,bang-etal-2023-multitask,qin-etal-2023-chatgpt}.
Therefore, following Section~\ref{sec:rl}, we not only utilize Recall@1, but we also leverage iEval to evaluate those methods on the test dataset as our evaluation metric.

For iEval, we set the number of simulated dialogue rounds to 3, and in the last round we let user simulator give the recommendations of the LLM a score of 0, 1, or 2.
To ensure the consistency in evaluating and scoring different models, we choose Llama3.1-8B as the external user simulator to evaluate the three models mentioned before.
Besides, we find that the evaluation results are very unstable when the number of test samples is less than 200, and gradually converge after the number larger than 400, so finally we choose 512 samples, considering the time consumption.
For Recall@1, we test them on all data in test dataset.

To validate the effectiveness and generalizability of our method for improving the multi-round conversation recommendation capability of large models, we conduct experiments on several well-known LLMs, Llama3.1-8B-instruct~\cite{dubey2024llama}, ChatGLM3-6B~\cite{glm2024chatglm} and Qwen2.5-7B-instruct~\cite{yang2024qwen2}.

\subsection{Main Results}

\noindent
\textbf{Performance Comparison}
Table~\ref{tab:compare} presents the Recall@1 and iEval results of different methods on the ReDial and OpenDialkg datasets.
Our method enhances the conversational recommendation potential of LLMs by RL training based on user simulator and employs a self-enhancement strategy to further improve performance during test time.
As shown in Table~\ref{tab:compare}, the proposed method, USB-Rec, outperforms all baselines on both datasets across most metrics. Specifically, USB-Rec achieves the highest iEval scores on both ReDial (1.29) and OpenDialKG (1.40), indicating superior recommendation quality. 
In terms of Recall@1, USB-Rec attains competitive performance, getting the sub-optimal result on ReDial dataset, the same as UniCRS (0.050), and achieving a near-top score on OpenDialKG (0.300), closely following BARCOR (0.312) and UniCRS (0.308).
Furthermore, compared to ReFICR, our method exhibits a marginal disadvantage in Recall@1 but demonstrates a significant improvement in iEval. 
This indicates that while RAG can enhance model accuracy within existing datasets, it may constrain the exploration of potentially superior recommendations.
These results demonstrate the effectiveness and robustness of USB-Rec across diverse conversational recommendation scenarios.



\noindent
\textbf{Generalizability Study}
In order to investigate the generalizability of our framework for training and inference, we evaluated the recommendation ability of three LLMs in Table~\ref{tab:two_dataset_comp_3llm}.
The results show that LLMs have gained significant improvement in recommendation ability after applying our framework.

Specifically, Llama has a certain recommendation potential on its own, so the value of the metric is significantly improved with SES applied to the Llama that was not fine-tuned.
Qwen and ChatGLM lacked the potential in conversational recommendation, and thus the effect of SES is slightly for them, being raised by 0.05 and 0.03 on average, respectively, without our fine-tuning framework.
However, after going through our training framework, SES can fully utilize their conversational recommendation capabilities, improving the value evaluation metric by more than 0.1, even if the gains are not evident after training alone, which are all less than 0.05.
This phenomenon illustrates that although the improvement of our fine-tuning method is not significant due to the noise in the dataset, the model could learn the potential of conversational recommendation ability, i.e., the output distribution of the model is approaching that of a recommendation expert, but the low probability makes it difficult to express.
And we demonstrate that our framework not only gives LLMs which lack conversation recommendation capabilities the potential to recommend, but brings this potential to full utilization.



\noindent
\textbf{Time-Consuming Comparison}
Since our strategy requires a lot of sampling and simulation interactions, we provide a time-consuming comparison with 8 H800 GPUs, as shown in Tabel~\ref{tab:time_ablation}.
Our algorithm increased the metric value by 0.08 using only 3s more time when it was applied only in the last round.
It is worth noting that while deeper tree search can better improve the performance of the model, the time consumption nearly becomes 9 times. If  parallel multiple API calls and vLLM acceleration are employed, the time increase caused by the tree search can be reduced to 3.67 seconds.

\begin{table}[tb]
\centering
\setlength{\tabcolsep}{7.8pt}
\caption{The time-consuming study of our method. The value time is the average time costed per sample.  The simulated turn is set to 3. T-S denotes tree search strategy.}
\begin{tabular}{ccccc}
\toprule
SES& T-S& Round& Average&  Time (s)  \\
 \midrule
\ding{55}          &-    & -              & 1.27& 2.93\\
 \midrule \noalign{\vskip 0.5ex}
\ding{51}          &\ding{55}     &Last 1             & 1.35& 6.12\\
\ding{51}          &\ding{55}     &Last 2               & 1.34& 8.70\\
\ding{51}          &\ding{51}    &Last 2                   & 1.37& 27.42\\
\bottomrule
\end{tabular}

\label{tab:time_ablation}
\end{table}

\subsection{Ablation Study}
In order to investigate the effect of each hyper-parameter of SES on the improvement of the recommendation ability of the large model, we choose Llama3.1-8b as a baseline to study and analyze.

\noindent
\textbf{Effect of Search Sampling Temperature}
We have conducted a detailed study of the temperatures of the LLMs responses with 3 simulation rounds, as shown in Figure~\ref{fig:ablation}a.
For the first response of the recommender LLM during the internal search, when the temperature is too low the LLM cannot explore more diversity with only 1.31 score, and when the temperature is too high the LLM tends to deviate from the goal bring the score down to 1.31. We finally found that 0.5 is a more appropriate value.

\noindent
\textbf{Effect of Majority Voting in SES}
Figure~\ref{fig:ablation}b shows that the influence of temperature is not very significant for the majority voting of internal user simulation, and the results are basically better when the temperature is high than when it is low. We chose 0.8 to avoid results that are too uncontrollable.
Furthermore, our experiments on the number of majority voting in Figure~\ref{fig:ablation}c show that increasing the number of times less than 10 improves the performance score from 1.35 to 1.38, and above 10 times it instead decreases the value of metric to 1.37, which is consistent with the phenomenon discussed in~\cite{chen2024more}.

\noindent
\textbf{Effect of Tree Search Depth in SES}
As shown in the Table~\ref{tab:ablation_sps_turn}, we have examined the rounds from which we should start our search strategy.
In order to show more comparisons, we set the number of simulated conversation rounds for evaluation to 4. 
From the last round to the last third round, the earlier the search strategy starts, the more significant the improvement of the recommendation ability of the LLM.
However, we find that the value of the metric drops slightly if the search strategy is used from the first round.
This is mainly due to the fact that the history is too short causing the user summarizer to not be able to summarize the user's preferences well, which leads to a large deviation between the subsequent internal user simulator's scoring and the external evaluation model's scoring misleading the LLM.
In addition, we also do ablation experiments on tree search, and we find that the randomness introduced by tree search can further increase the credibility of internal scoring on top of the majority voting mechanism.
\begin{table}[tb]
\caption{The ablation study of when to begin SES and the effect of tree search accordingly. The simulated number of rounds is set to 4. T-S represent the Tree Search method.}
\centering
\setlength{\tabcolsep}{3.4pt}
\begin{tabular}{cccccc}
\toprule
       \,SES\, & Round & \,T-S\, & \,ReDial\, & OpenDialkg & Average  \\ 
\midrule
        \ding{55}& - & -& 1.30& 1.41& 1.36\\ 
\midrule \noalign{\vskip 0.5ex}
        \ding{51} & Last 1 & -& 1.36& 1.45& 1.41\\   
        \ding{51} & Last 2 & \ding{55} & \underline{1.40}& 1.48& 1.44\\ 
        \ding{51} & Last 3 & \ding{55} & 1.39& 1.46& 1.43\\ 
        \ding{51} & Last 4 & \ding{55} & 1.35& 1.44& 1.40\\ 
\midrule \noalign{\vskip 0.5ex}
        \ding{51} & Last 2 & \ding{51} & \textbf{1.43}& \textbf{1.50}& \textbf{1.47}\\     
        \ding{51} & Last 3 & \ding{51} & \underline{1.40}& \underline{1.49}& \underline{1.45}\\ 
        \ding{51} & Last 4 & \ding{51} & 1.37& 1.46& 1.42\\ 
\bottomrule
\end{tabular}

\label{tab:ablation_sps_turn}
\end{table}

\begin{table}[]
\caption{The ablation study of the number of responses sampled in the first round of SES and the number of simulated rounds. The SES is only applied in the last round.}
\centering
\setlength{\tabcolsep}{2.8pt}
\begin{tabular}{ccccc}
\toprule
 \# of Turns & \# of Resp. & ReDial& OpenDialkg & Average   \\
 \midrule
\multirow{4}{*}{3}              & 2              & 1.26& 1.36& 1.31\\
               & 3              & 1.29& 1.40& 1.35\\
               & 4              & 1.28& 1.39& 1.34\\
               & 5              & 1.24& 1.32& 1.28\\
\midrule \noalign{\vskip 0.5ex}
 \multirow{4}{*}{4}               & 2              & 1.33& 1.43& 1.38\\
               & 3              & 1.36& 1.45& 1.41\\
               & 4              & 1.35& 1.44& 1.40\\
               & 5              & 1.32& 1.42& 1.37\\
\midrule \noalign{\vskip 0.5ex}
 \multirow{4}{*}{5}               & 2              & 1.35& \underline{1.46}& 1.40\\
               & 3              & \textbf{1.38}& \textbf{1.47}& \textbf{1.43}\\
               & 4              & \underline{1.37}& \textbf{1.47}& \underline{1.42}\\
               & 5              & 1.33& 1.44& 1.39\\
\bottomrule
\end{tabular}

\label{tab:ablation_sps_turn_resp}
\end{table}

\begin{table}[t]
\caption{
An comparison example of Llama3.1-8B and our method
}
\centering
\scalebox{0.80}{
\begin{tabular}{P{10cm}}
\toprule[1pt]
\textbf{Label: }  \colorbox {gray!40}{Zero Dark Thirty}\\
\midrule
\textbf{Sampled History Seed} \\
\midrule \noalign{\vskip 0.5ex}
\textbf{User:} I like Black Hawk Down. I saw it the other night for the tenth time. Do you know of any other films that are somewhat similar?\\
\textbf{Human Recommender:} Black Hawk Down is an action film and a war film is that why you like it?\\
\textbf{User:} I think I like action more, but I do enjoy this type of war film. Could you give me examples of similar war films?\\
\midrule
\textbf{Simulated Conversation for Evaluation --- Llama3.1-8B} \\
\midrule \noalign{\vskip 0.5ex}
\textbf{CRS:} You might enjoy Platoon, We Were Soldiers, and Saving Private Ryan, all intense war dramas. Have you seen any of these?\\
\textbf{User Simulator:} No, I haven't seen those. They sound like they're about the US military in combat situations, with a strong focus on realism.\\
\textbf{CRS:} Those films are gritty and realistic, like Black Hawk Down. Do you prefer historical or fictional war stories?\\
\textbf{User Simulator:} I prefer historical war stories, especially those based on true events and real people. They tend to be more intense and suspenseful.\\
Please explain your last time of recommendation.\\
\textbf{CRS:} Based on your preference for historical war stories, I recommended \colorbox {gray!40}{Platoon}, \colorbox {gray!40}{We Were Soldiers}, and \colorbox {gray!40}{Saving Private Ryan}. These films are all based on true events and are known for their intense and realistic portrayals of war. My previous recommendation was based on your preference for a courtroom drama with a strong focus on character development and social commentary, and your desire for an intense and dramatic tone.\\
Does the explanation make you want to accept the recommendation? Please give your score ...\\
\midrule \noalign{\vskip 0.5ex}
\textbf{Score: }1\\
\midrule
\textbf{Simulated Conversation for Evaluation --- Llama3.1-8B with RL \& SES} \\
\midrule \noalign{\vskip 0.5ex}
\textbf{CRS:} You enjoy intense action and war themes. Have you seen Saving Private Ryan or We Were Soldiers?\\
\textbf{User Simulator:} Those are war films, but I'm looking for something more recent and intense.\\
\textbf{CRS:} You're looking for a recent, intense war film based on a real-life operation. Lone Survivor (2013) comes to mind, based on a true Navy SEAL story.\\
\textbf{User Simulator:} 
It's a biographical war film, but I'm looking for something more focused on a specific mission. It's a thriller that's based on a true story, with a strong focus on the hunt for a high-value target.\\
Please explain your last time of recommendation\\
\textbf{CRS:} My last recommendation, Lone Survivor, is a biographical war film, not a specific mission-based thriller. I'll try again. I'll recommend  \colorbox {gray!40}{Zero Dark Thirty (2012)}, a thriller based on a true story, focusing on the hunt for Osama bin Laden.
\\Does the explanation make you want to accept the recommendation? Please give your score ...\\
\midrule \noalign{\vskip 0.5ex}
\textbf{Score: }2\\
\bottomrule[1pt]

\end{tabular}
}
\label{case:llama_org}
\end{table}

\noindent
\textbf{Effect of Tree Search Width in SES}
In Table~\ref{tab:ablation_sps_turn_resp}, we examine the effect on performance of different number of simulation rounds and the number of response samples taken in the first round of SES, with fixed number of majority voting, 10.
As the number of simulation rounds increases, the LLMs' recommendation become more effective. The performance is improved noticeably as the number of simulated rounds increases from 3 to 4, with an average increase of evaluation metric about 0.06, however, after 4 rounds the improvement is gradually insignificant, with an average gain of only 0.02 from 4 to 5 rounds.
This means that more and deeper interactions can provide more valid information for LLMs to make better recommendations, however there is also an upper limit to how much that can help the LLMs, which is constrained by the summarization and reasoning ability of the LLMs themselves for we choose a small version of those LLMs due to the limited computational resources.

What's more, we find that the performance does not get better the more times the responses are sampled. Instead, the metric values start to decline when the number of sampling is more than 4.
The results from 5 sampling times are on average 0.02 or 0.03 lower than those from 2 sampling times, and an average of 0.04 to 0.06 lower than those from 3 sampling times.
This is mainly due to the fact that the internal model doesn't know the real labels, and when faced with too many choices, it can easily be misled, leading to deteriorating performance.

\subsection{Case Study}
As shown in Table~\ref{case:llama_org}, we present a comparison example using the Llama3.1-8B as a case study to demonstrate the effects of incorporating our proposed methodology. 
The case demonstrates that the baseline model persistently recommend the films they initially deem appropriate throughout the interaction process, without adapting to user's feedback or generating revised suggestions.
While the base model with our methodology demonstrates adaptive recommendation dynamics, where although initial suggestions mirror baseline Llama outputs, it progressively refines recommendations through iterative feedback analysis, and ultimately achieving the user preference objectives.

\section{CONCLUSION}

In this work, we propose USB-Rec, a training-inference integrated framework for improving LLMs' conversational recommendation capability at the model level.
At training stage, an automatic preference pair dataset construction method is proposed for RL training.
At inference stage, We design a search strategy that can deeply explore the conversational recommendation capabilities of LLMs that acquired from training stage but not be fully expressed.
Extensive experiment results show that USB-Rec outperforms previous traditional NLP-model-based and LLM-based methods, demonstrating its effectiveness and the generalizability for different base LLMs.


\bibliographystyle{ACM-Reference-Format}
\bibliography{sample-base}

\end{document}